\documentclass[conference, a4paper]{IEEEtran}
\IEEEoverridecommandlockouts
\usepackage{balance}
\usepackage{cite}
\usepackage{graphicx}
\usepackage{lipsum}
\graphicspath{{Figures/}}
\usepackage{caption}
\usepackage{subfig} 
\usepackage{color}
\usepackage{balance}
\usepackage{float}
\usepackage[hyphens,spaces,obeyspaces]{url}
\urldef{\mycaseurl}\url{https://www.adaptcentre.ie/case-studies/ai-based-advert-creation-system-for-next-generation-publicity}
\usepackage{float}
\usepackage{breqn}
\def\BibTeX{{\rm B\kern-.05em{\sc i\kern-.025em b}\kern-.08em
    T\kern-.1667em\lower.7ex\hbox{E}\kern-.125emX}}

\usepackage{blindtext}

\usepackage{tikz}
\usetikzlibrary{positioning}

\begin{document}

\title{The ALOS Dataset for Advert Localization \\in Outdoor Scenes}

\author{Soumyabrata~Dev, Murhaf~Hossari, Matthew~Nicholson, Killian~McCabe, Atul~Nautiyal\\
Clare~Conran, Jian~Tang, Wei~Xu, and~Fran\c{c}ois~Piti\'e
\thanks{The ADAPT Centre for Digital Content Technology is funded under the SFI Research Centres Programme (Grant 13/RC/2106) and is co-funded under the European Regional Development Fund.}        
\thanks{S.\ Dev, M.\ Hossari, M.\ Nicholson, K.\ McCabe, A.\ Nautiyal, C.\ Conran,  and F.\ Piti\'e are with the ADAPT SFI Research Centre, Trinity College Dublin. F.\ Piti\'e is also with the Department of Electronic \& Electrical Engineering, Trinity College Dublin.}
\thanks{J.\ Tang and W.\ Xu are with the Huawei Ireland Research Center, Dublin.}
\thanks{Send correspondence to F.\ Piti\'e (PITIEF@tcd.ie).}
}

\IEEEoverridecommandlockouts
\IEEEpubid{\makebox[\columnwidth]{978-1-5386-8212-8/19/\$31.00 \copyright 2019 IEEE \hfill} \hspace{\columnsep}\makebox[\columnwidth]{ }}

\maketitle

\begin{abstract}
The rapid increase in the number of online videos provides the marketing and advertising agents ample opportunities to reach out to their audience. One of the most widely used strategies is product placement, or embedded marketing, wherein new advertisements are integrated seamlessly into existing advertisements in videos. Such strategies involve accurately localizing the position of the advert in the image frame, either manually in the video editing phase, or by using machine learning frameworks. However, these machine learning techniques and deep neural networks need a massive amount of data for training. In this paper, we propose and release the first large-scale dataset of advertisement billboards, captured in outdoor scenes. We also benchmark several state-of-the-art semantic segmentation algorithms on our proposed dataset.
\end{abstract}


\begin{tikzpicture}[overlay, remember picture]
  \path (current page.north) node (anchor) {};
  \node [below=of anchor] {%
  2019 Eleventh International Conference on Quality of Multimedia Experience (QoMEX)};
\end{tikzpicture}

\section{Introduction}

With the advent of rapid growth in internet services and the increase in the number of online viewers, there has been a massive increase in the number of online videos. Now-a-days, the marketing strategies involve product placement, wherein new adverts are seamlessly integrated into the original videos~\cite{nautiyal2018advert,hossari2018adnet}.
This provides opportunities for the advertisement and marketing agencies to reach out to the diverse audience via advertisements. During the post-processing stage of the video, the editors replace the existing advert object in the scene, with a new advert. 

With the advent of high computing powers, there has been a massive advancement in the field of computer vision and image processing. The task of object recognition and object classification has become easier for the past decade -- thanks to the release of large-scale datasets and their associated challenges. In order to train the machine-learning algorithms, it is important to have a large-scale dataset~\cite{dev2019case}, along with manually annotated class labels. Currently, there are no available datasets of billboard advertisement images, along with manually annotated labels of advertisement position.

The main contributions of this paper are as follows: (a) we propose and release the first large-scale dataset of billboard images, along with high quality annotations of billboard location, (b) we also provide a detailed and systematic evaluation of popular segmentation algorithms on our proposed dataset. In this paper, we restrict our domain to outdoor scene images, that are mostly captured using car dashboard cameras. We collaborated with Mapillary~\footnote{\url{https://www.mapillary.com/}} -- a crowd-sourcing service for sharing geotagged photos, and is releasing a large dataset of outdoor scene images together with manually annotated labels. We believe that this dataset will help researchers in building state-of-the-art algorithms for efficient advert detection.

\begin{figure*}[htb]
\centering

\includegraphics[height=0.11\textwidth]{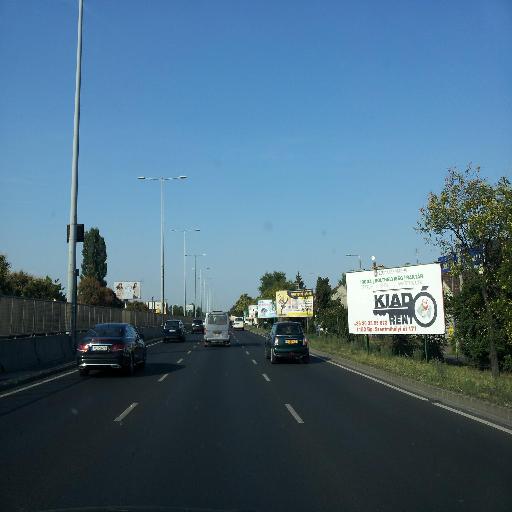}
\includegraphics[height=0.11\textwidth]{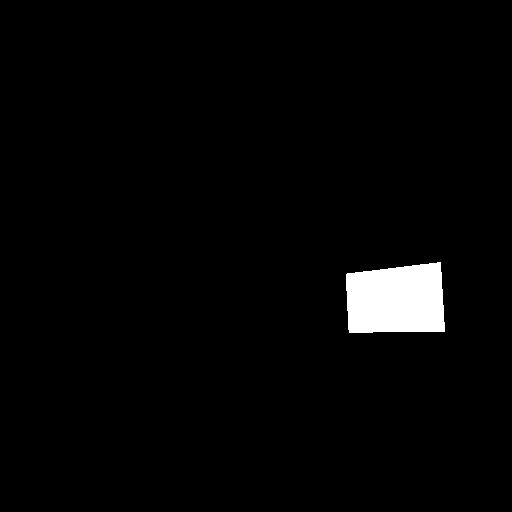}
\includegraphics[height=0.11\textwidth]{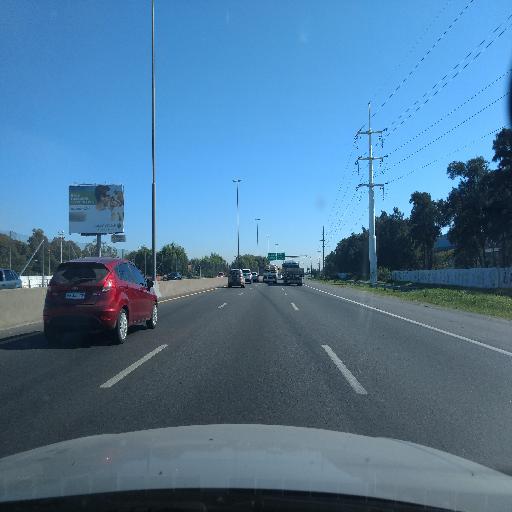}
\includegraphics[height=0.11\textwidth]{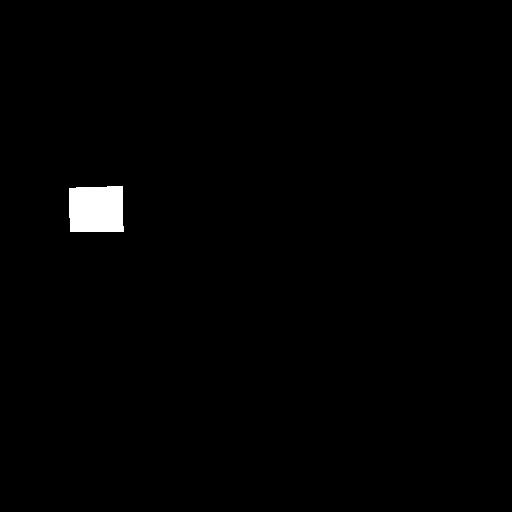}
\includegraphics[height=0.11\textwidth]{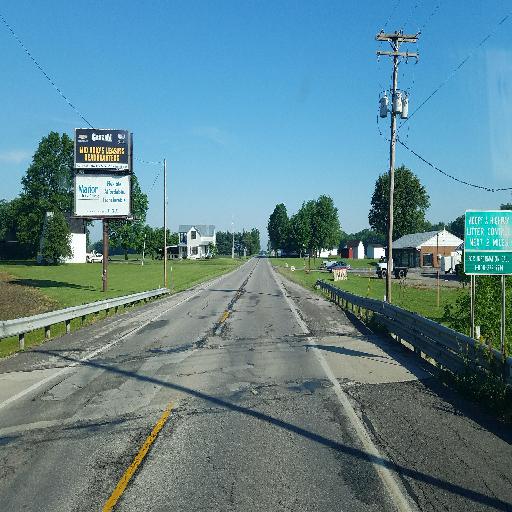}
\includegraphics[height=0.11\textwidth]{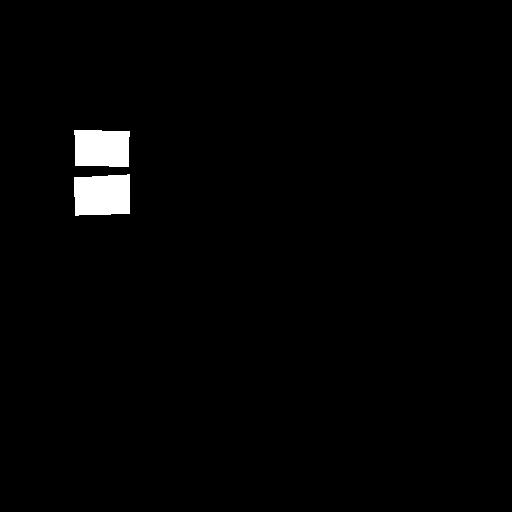}
\includegraphics[height=0.11\textwidth]{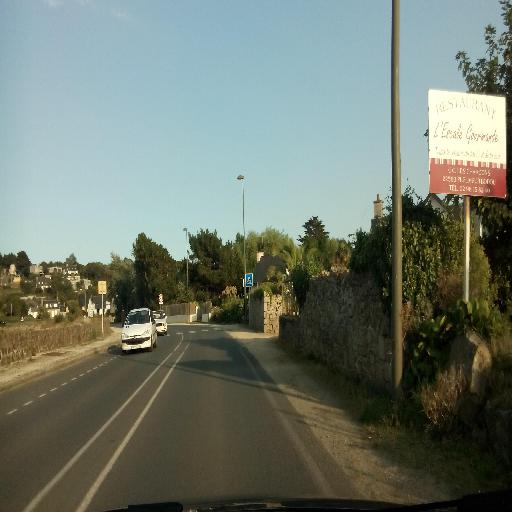}
\includegraphics[height=0.11\textwidth]{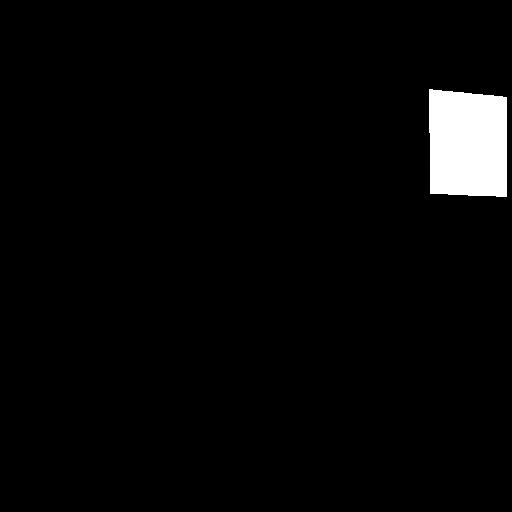}\\
\includegraphics[height=0.11\textwidth]{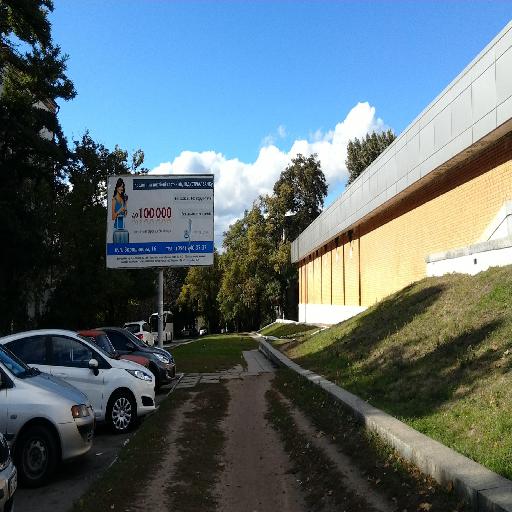}
\includegraphics[height=0.11\textwidth]{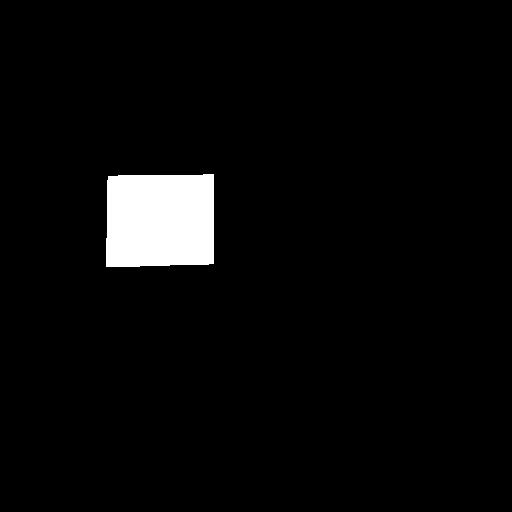}
\includegraphics[height=0.11\textwidth]{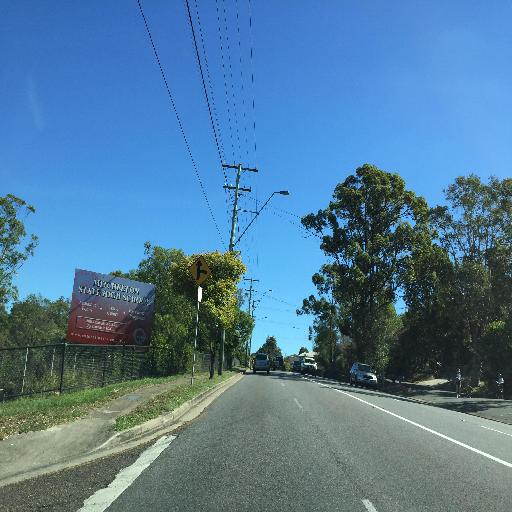}
\includegraphics[height=0.11\textwidth]{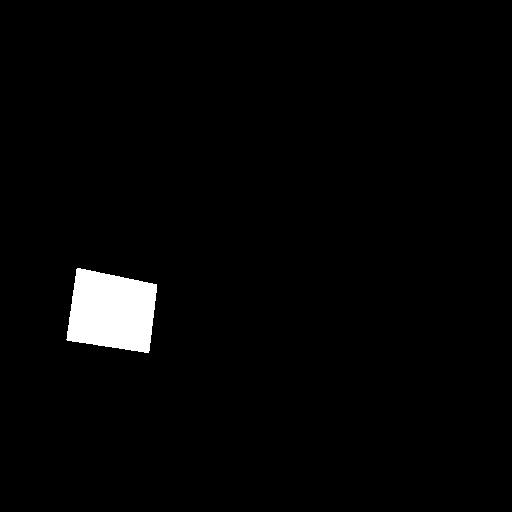}
\includegraphics[height=0.11\textwidth]{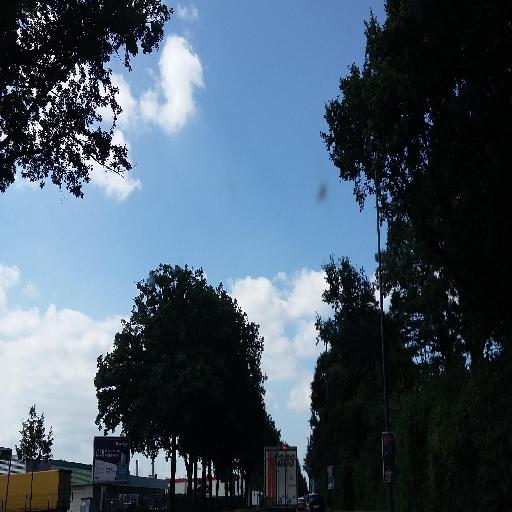}
\includegraphics[height=0.11\textwidth]{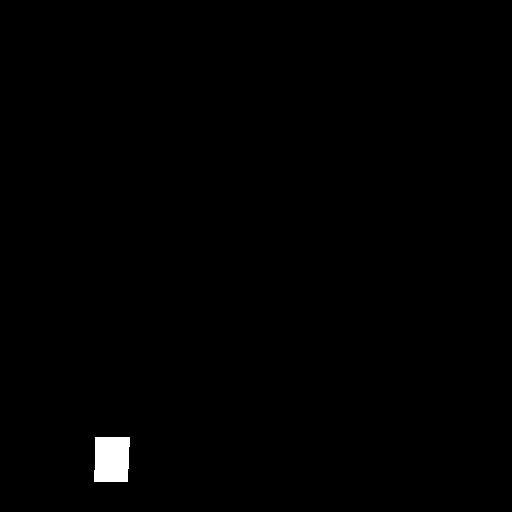}
\includegraphics[height=0.11\textwidth]{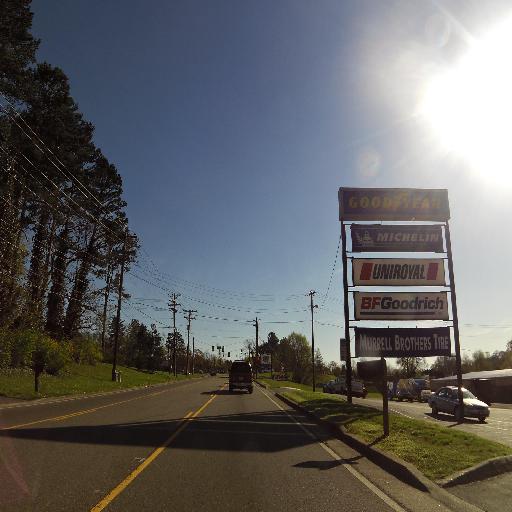}
\includegraphics[height=0.11\textwidth]{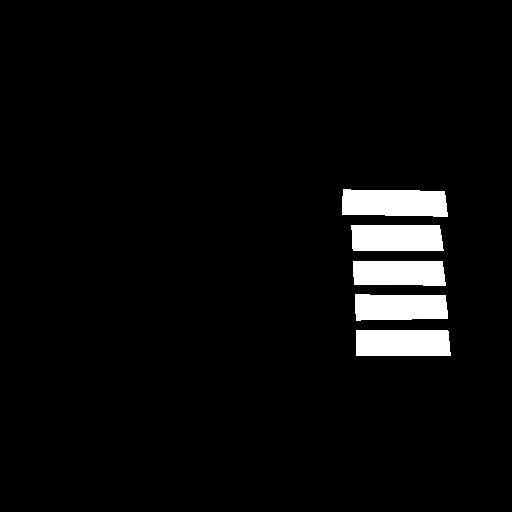}
\caption{Representative images of the ALOS dataset, along with manually annotated binary image maps.}
\label{fig:sample-images}
\end{figure*}

\section{Dataset}
\label{sec:data-all}

\subsection{Dataset organization}
We refer to our billboard dataset as the \textbf{ALOS} dataset, that stands for \textbf{A}dvert \textbf{L}ocalization in \textbf{O}utdoor \textbf{S}cenes. The ALOS dataset contains images from Mapillary geotagged images. Figure~\ref{fig:sample-images} shows a few representative images, along with its manually annotated ground-truth labels.

Mapillary is a crowd-sourcing platform, that provides geotagged images, mostly street view style images from dashboard cameras. They have licensable images that can be filtered to include only images that they have detected which contain billboards. Mapillary have a platform where users can share geotagged images. Their community can then edit/label/review uploaded images. These images are from around the world, captured via various types camera. The Mapillary platform has the ability to filter the images to ones that contain billboards. The definition of billboard, as per Mapillary platform, is more broad than the target of our work. Therefore, the filtered images will contain images consisting of some signage or shop front, which should be neglected in our case. Therefore, it is absolutely necessary for us to annotate the images ourselves, ensuring the four corners are accurately labeled. Such high-quality annotated labels can subsequently used in several use cases. 

The dataset contain billboards, posters and screens that are potential candidates for advertisement placement. In this paper, we interchangeably use the terms \emph{advert} and \emph{billboard} to represent a candidate object for advertising. A general rule of the image selection in the dataset includes good contrast, legible text and good image resolution (at least $800\times600$ pixels). We restrict the shape of the billboard to any four-sided convex polygon. Moreover, the amount of perspective distortion should be at the minimum. A frontal viewpoint of the billboard picture is the ideal. The maximum allowable perspective distortion is $15$ degrees. Moreover, the billboard should not be occluded to a great extent. It should be visible with less (or absolutely no) additional human effort. The amount of occlusion should not be more than 10\% of the billboard image. Finally, the billboard should not cover most of the entire image dimension.

Based on these restrictions, we build a corpus~\footnote{The download link of the dataset is available here: \mycaseurl} of images that are suitable for the task of billboard localization in imagze frames. The total number of images in this dataset is $8065$.

\subsection{Dataset characteristics}
\begin{figure}[htb]
\centering
\subfloat[Billboard coverage]{\includegraphics[height=0.16\textwidth]{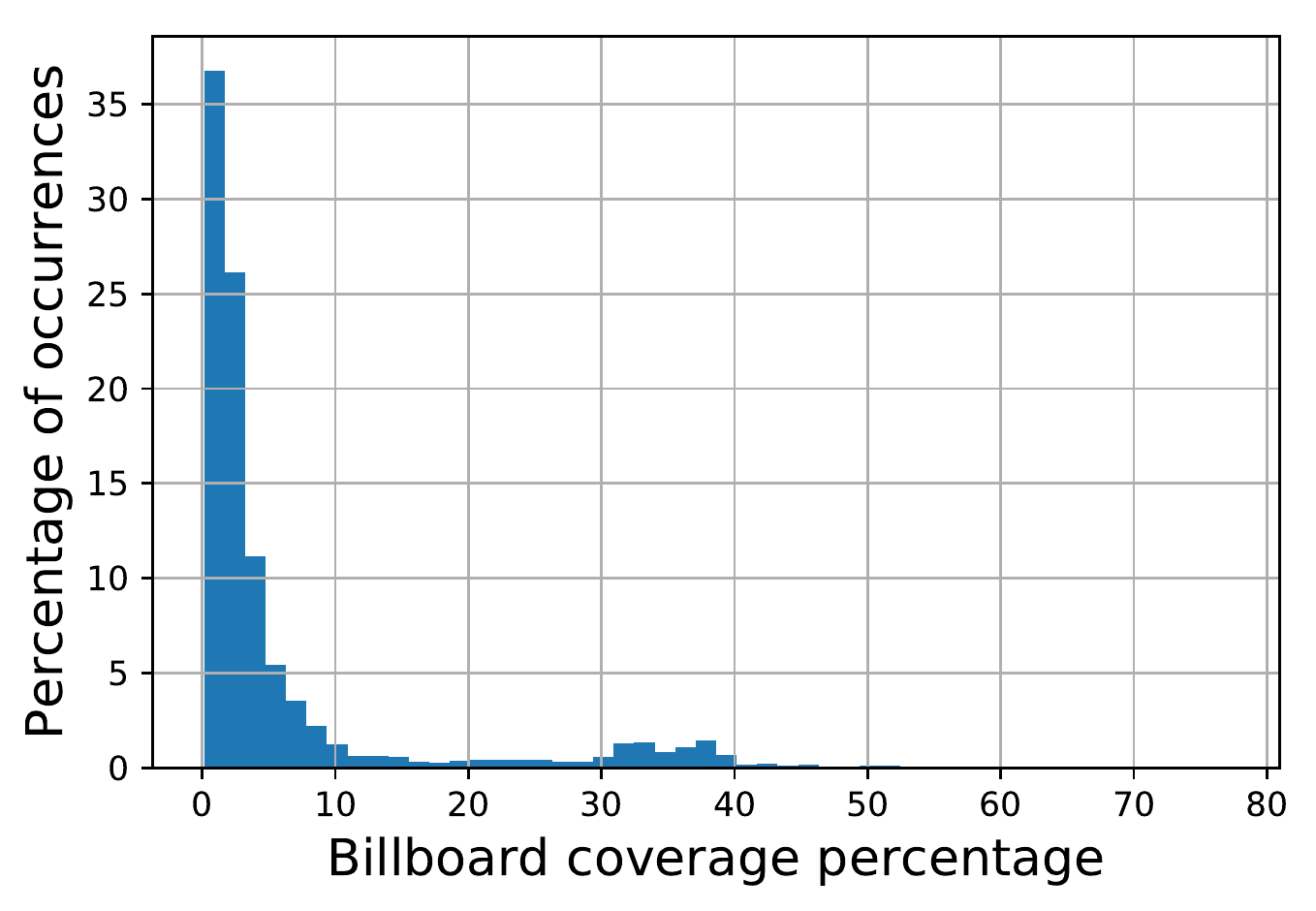}}\ 
\subfloat[Number of billboards]{\includegraphics[height=0.16\textwidth]{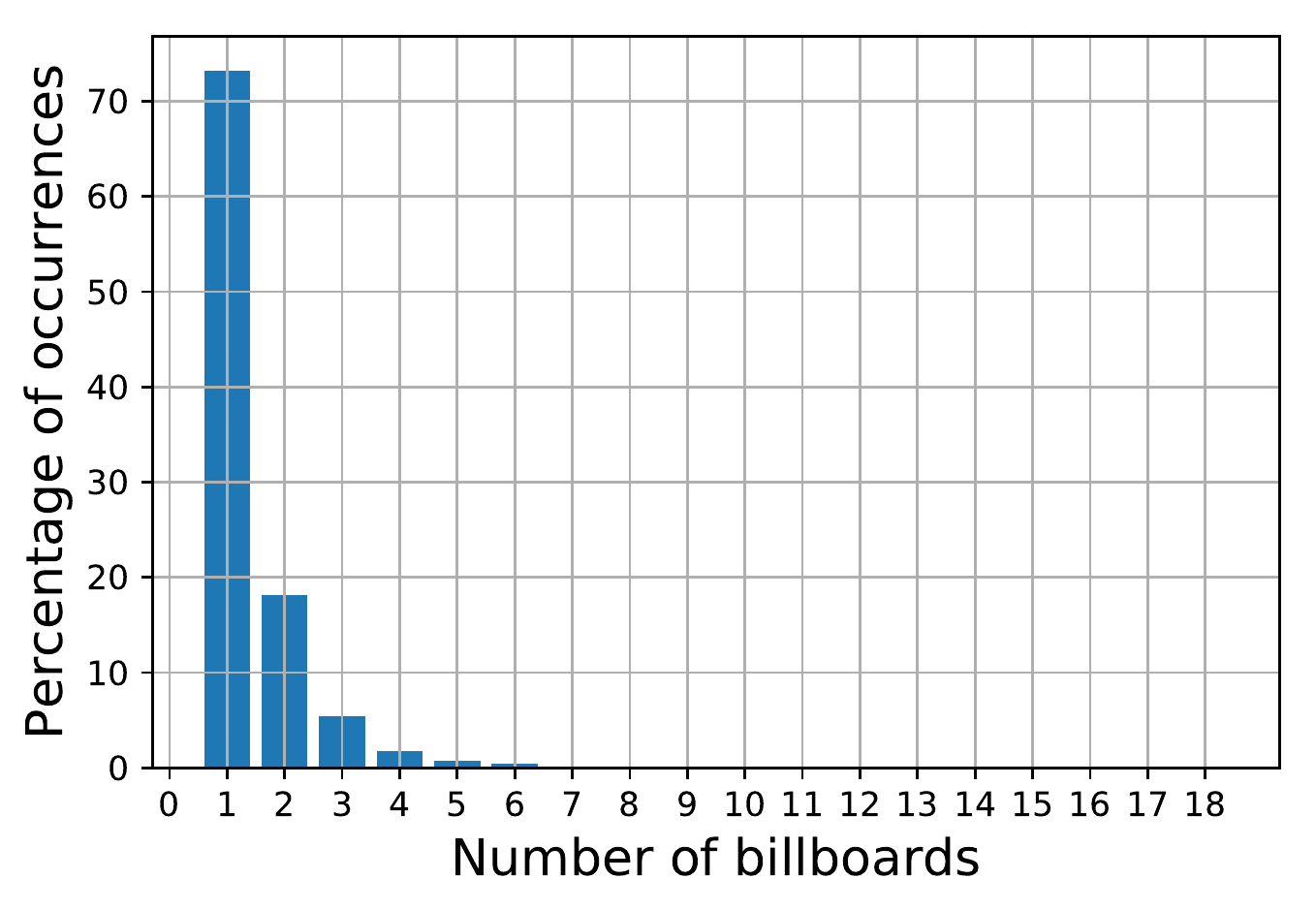}}\\
\subfloat[State of occlusion]{\includegraphics[height=0.16\textwidth]{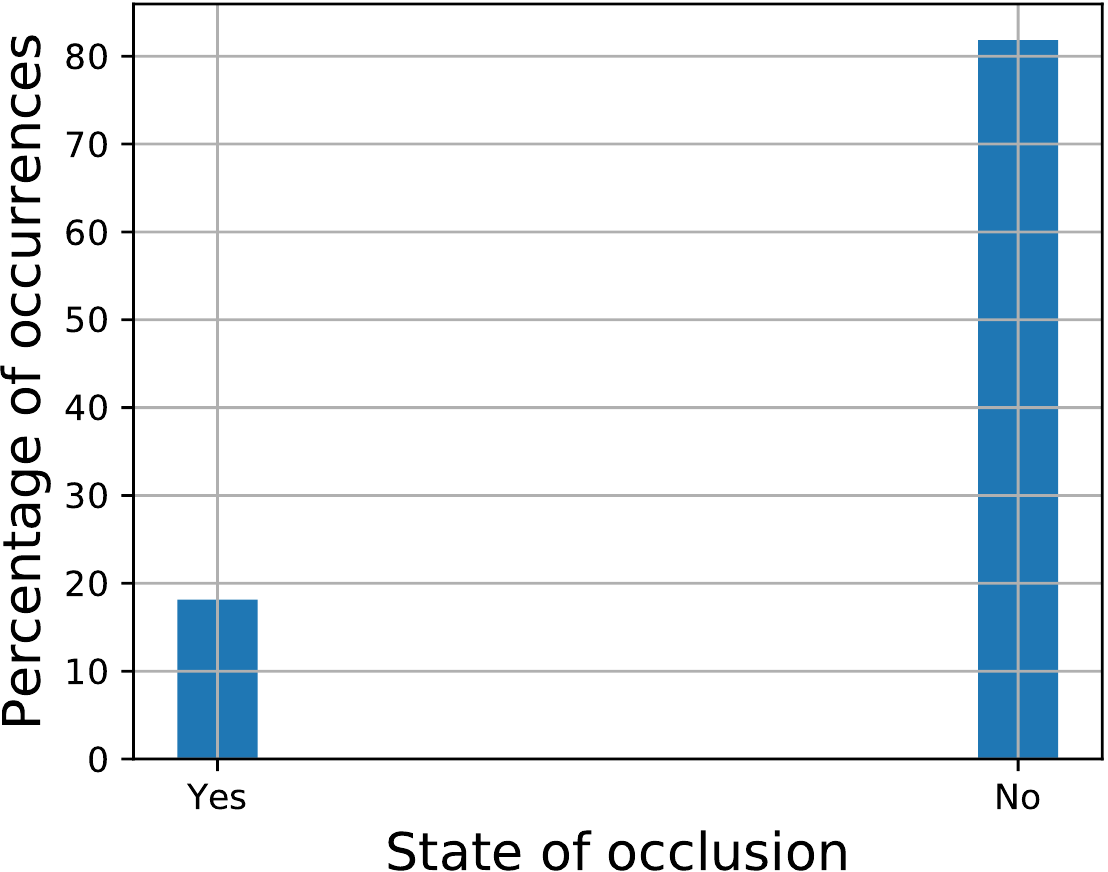}}\ 
\subfloat[State of off-screen]{\includegraphics[height=0.16\textwidth]{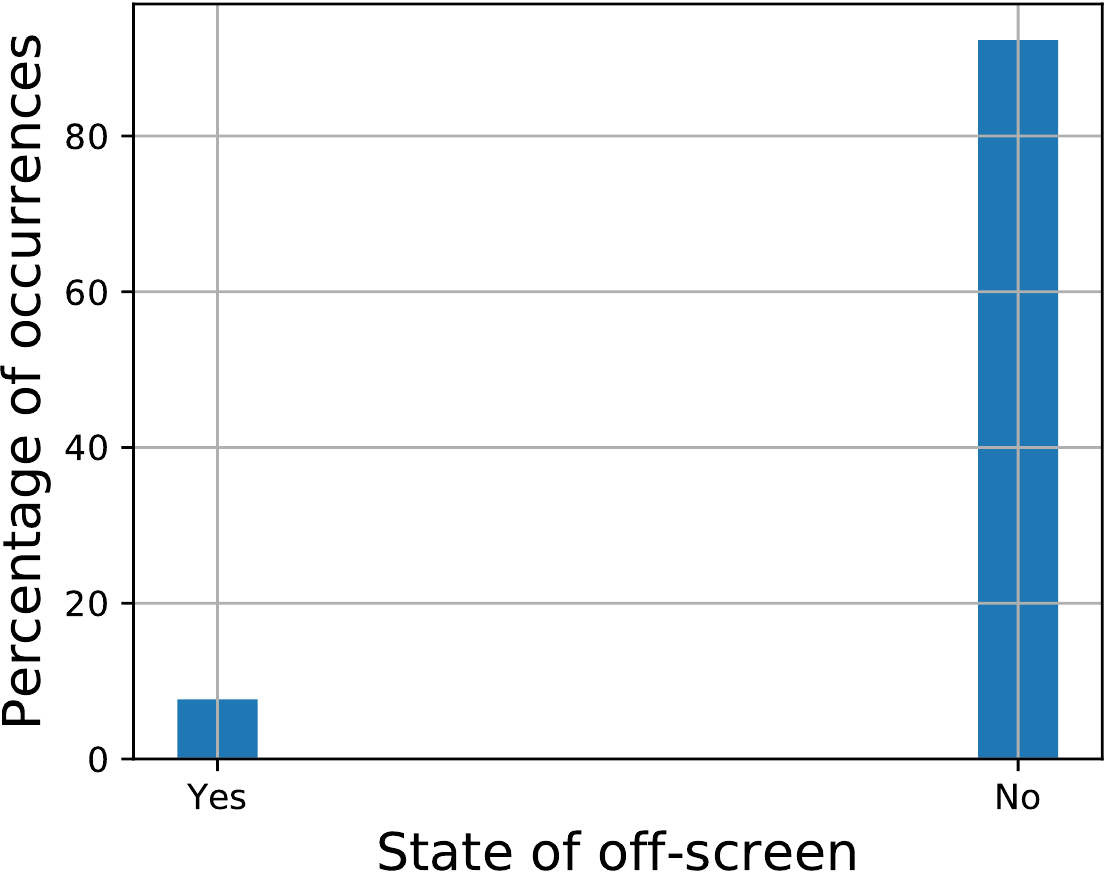}}
\caption{We visualize the distributions w.r.t.\ (a) billboard coverage area in a single image, (b) number of billboards in a single image, (c) state of occlusion, and (d) state of off-screen.}
\label{fig:dataset-char}
\end{figure}

During the curation of the ALOS dataset, we ensured that diverse characteristics of the billboards are included in the dataset. The billboards in the ALOS dataset cover varying proportion of the total image area. We refer the amount of area covered by billboard as billboard coverage. The smallest billboard in the ALOS dataset cover $0.13\%$ of the image area; while the largest billboard capture $77.08\%$ of its image area. We show the distribution of the billboard coverage in Fig.~\ref{fig:dataset-char}(a). Similarly, the number of billboards in a single image varies significantly across the images of the ALOS dataset. Most of the images have a single billboard (cf.\ Fig.~\ref{fig:dataset-char}(b)) in it, while the largest number of billboards in a single image is $18$. Finally, we also include images in our dataset, whose billboards are partially covered with occlusions (cf.\ Fig.~\ref{fig:dataset-char}(c)), or in the state of the being off-screen (cf.\ Fig.~\ref{fig:dataset-char}(d)). Figure~\ref{fig:dataset-char} summarizes the distribution of billboard coverage and number of billboards in the proposed dataset.

\section{Benchmarking Experiments}
\label{sec:exps}
We implement three deep-learning models for semantic segmentation on our dataset: Fully Convolutional Network (FCN)~\cite{long2015fully}, Pyramid Scene Parsing Network (PSPNet)~\cite{zhao2017pyramid}, and U-Net~\cite{ronneberger2015u}. In the domain of visual computing, convolutional neural networks have shown promising results in dense semantic segmentation of images. 

Long et al.\ have shown that a fully convolutional network trained end-to-end, pixel-by-pixel, can produce detailed segmentation maps of input images~\cite{long2015fully}. Recently, Zhao et al.\ proposed the pyramid scene parsing network~\cite{zhao2017pyramid} that produced the best results on ImageNet scene parsing challenge 2016. It uses various region-based context aggregation, using its pyramid pooling module. We also benchmark U-Net architecture~\cite{ronneberger2015u} on our proposed dataset. 

\subsection{Subjective Evaluation}
\begin{figure}[htb]
\centering
\includegraphics[height=0.092\textwidth]{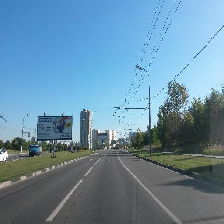}
\includegraphics[height=0.093\textwidth]{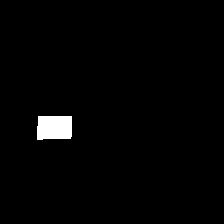}
\includegraphics[height=0.092\textwidth]{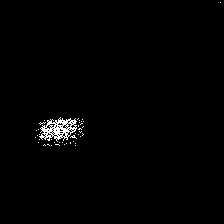}
\includegraphics[height=0.092\textwidth]{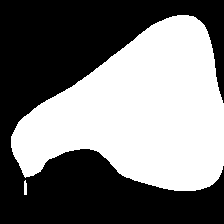}
\includegraphics[height=0.092\textwidth]{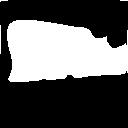}\\
\includegraphics[height=0.092\textwidth]{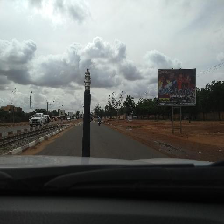}
\includegraphics[height=0.093\textwidth]{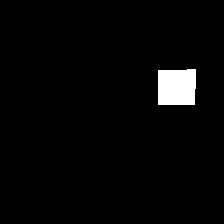}
\includegraphics[height=0.092\textwidth]{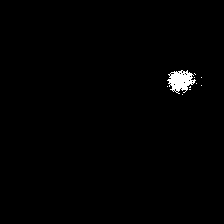}
\includegraphics[height=0.092\textwidth]{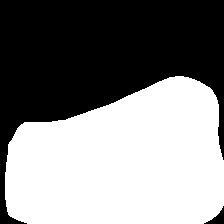}
\includegraphics[height=0.092\textwidth]{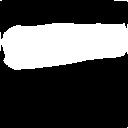}\\
\includegraphics[height=0.092\textwidth]{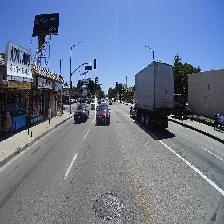}
\includegraphics[height=0.093\textwidth]{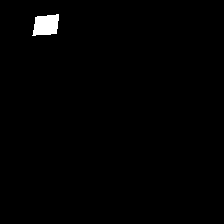}
\includegraphics[height=0.092\textwidth]{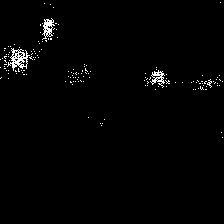}
\includegraphics[height=0.092\textwidth]{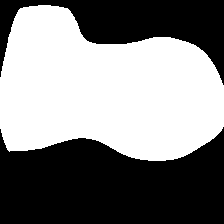}
\includegraphics[height=0.092\textwidth]{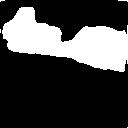}\\
\includegraphics[height=0.092\textwidth]{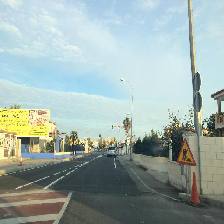}
\includegraphics[height=0.093\textwidth]{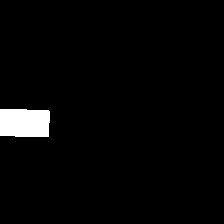}
\includegraphics[height=0.092\textwidth]{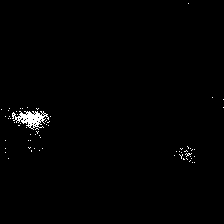}
\includegraphics[height=0.092\textwidth]{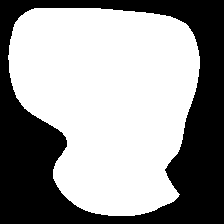}
\includegraphics[height=0.092\textwidth]{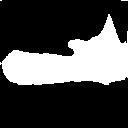}\\
\vspace{-0.3cm}
\subfloat[\scriptsize{Input image}]{\includegraphics[height=0.092\textwidth]{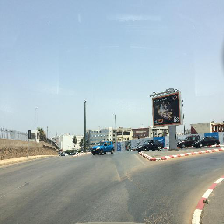}}\ 
\subfloat[\scriptsize{Ground truth}]{\includegraphics[height=0.093\textwidth]{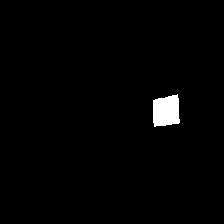}}\ 
\subfloat[\scriptsize{FCN}]{\includegraphics[height=0.092\textwidth]{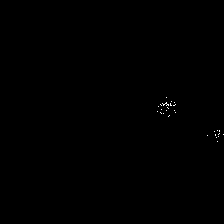}}\ 
\subfloat[\scriptsize{PSPNet}]{\includegraphics[height=0.092\textwidth]{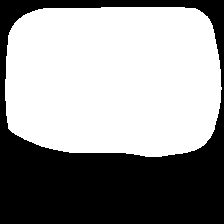}}\ 
\subfloat[\scriptsize{U-Net}]{\includegraphics[height=0.092\textwidth]{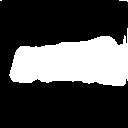}}
\caption{Subjective evaluation of various semantic segmentation algorithms in ALOS dataset. We show the (a) input image, (b) corresponding binary ground-truth images, and results obtained from (c) FCN, (d) PSPNet, (e) U-Net.}
\label{fig:subj-eval}
\end{figure}

We train the FCN network on resized images of size $256\times256$ of the ALOS dataset. We train it for $63000$ iterations, with a batch size of $2$ images. A smaller batch size is necessary to fit the entire model in memory during the training time. We save the model at the end of training, for evaluation purposes. We train the PSPNet model for $500$ epochs, with a batch size of $2$ images. Similar to FCN network, we use a small batch size so that the entire network can fit in the memory during training. Finally, we trained U-Net model on the ALOS dataset. We used Adam optimiser, and trained the model for $10000$ steps. 
Figure~\ref{fig:subj-eval} shows a few sample visual results of the benchmarking algorithms on the ALOS dataset. The results from PSPNet model are poor, as the model fails to converge properly due to its large network size. The FCN network, being the simplest among the benchmarking methods generalizes well, and works well in most of the cases. 

\subsection{Objective Evaluation}
In addition to the subjective evaluation, we also provide the average values of several objective metric, that are commonly used in the area of semantic segmentation. Suppose $n_{ij}$ is the number of pixels of class $i$, that are predicted to class $j$. We define $n_{cl}$ as the total number of classes in the task of semantic segmentation. The total number of pixels in class $i$ is defined as $t_i = \sum_{j=1}^{n_{cl}} n_{ij}$. We compute several metrics for the subjective evaluation of the different approaches. The pixel accuracy is defined by $\frac{\sum_{i}^{} n_{ii}}{\sum_{i}^{} t_{i}}$. The mean accuracy is defined as $\frac{1}{n_{cl}}\sum_{i}^{}\frac{n_{ii}}{t_i}$. The mean intersection over union is defined by $\frac{1}{n_{cl}}\frac{\sum_{i}^{}n_{ii}}{t_i+\sum_{j}^{}n_{ji}-n_{ii}}$. Finally, the frequency weighted intersection over union is defined as $\frac{1}{\sum_{k}^{}t_k}\frac{\sum_{i}^{}t_in_{ii}}{t_i+\sum_{j}^{}n_{ji}-n_{ii}}$.

We compute the average values of pixel accuracy, mean accuracy, mean intersection over union, and frequency weighted intersection over union, across all the images of the proposed dataset. Table~\ref{table:result} summarizes the results of the various algorithms. 

\begin{table}[htb]
\centering
\begin{tabular}{l|p{1.3cm}|p{1.3cm}|p{1.3cm}|p{1.3cm}}
       & Pixel Accuracy & Mean Accuracy & Mean IOU & Frequency Weighted IOU \\ \hline
FCN    & \textbf{0.962}  & 0.699  &   \textbf{0.638}   &  \textbf{0.937} \\ \hline
PSPNet & 0.554 & 0.558 & 0.304 & 0.521 \\ \hline
U-Net &  0.721  &  \textbf{0.814}  &  0.432 & 0.689 \\ \hline 
\end{tabular}
\caption{Benchmarking of the ALOS dataset with various deep-learning based segmentation algorithms. The best performance according to each metric is marked in bold.}
\label{table:result}
\end{table}

\subsection{Discussion on Benchmarking}

These benchmarking results provide us interesting insights. We observe that a light-weight FCN network performs the best across the various methods. This is mainly because the larger network fail to converge on our dataset. The results of the PSPNet can be further improved by using batch normalization across multiple graphics processing units (GPUs). The U-Net approach performs the best based on mean accuracy, but performs poor according to other important metrics of semantic segmentation. Therefore, we conclude that the overall scores could be further improved by proposing a bespoke shallower neural network, that is specifically tailored for advert localization in outdoor scenes.

\section{Conclusions and Future Works}
\label{sec:conc}
In this paper, we propose and release ALOS -- a large-scale dataset of billboard/advert images, along with high-quality manually annotated ground-truth images. This dataset will be particularly useful for advertising and marketing agencies, for the purpose of product placement and embedded marketing. We also benchmark the performance of several popular deep-learning based segmentation algorithms in our proposed dataset. In the future, we plan in relaxing the criterion of outdoor scenes, and propose a larger dataset that encompasses other domains, including indoor scenes and entertainment videos.

\balance

\end{document}